# Progressive Cross Attention Network for Flood Segmentation using Multispectral Satellite Imagery

Vicky Feliren, Fithrothul Khikmah, Irfan Dwiki Bhaswara, Bahrul I. Nasution, Alex M. Lechner, Muhamad Risqi U. Saputra*

*Abstract*—In recent years, the integration of deep learning techniques with remote sensing technology has revolutionized the way natural hazards, such as floods, are monitored and managed. However, existing methods for flood segmentation using remote sensing data often overlook the utility of correlative features among multispectral satellite information. In this study, we introduce a progressive cross attention network (ProCANet), a deep learning model that progressively applies both self- and cross-attention mechanisms to multispectral features, generating optimal feature combinations for flood segmentation. The proposed model was compared with state-of-the-art approaches using Sen1Floods11 dataset and our bespoke flood data generated for the Citarum River basin, Indonesia. Our model demonstrated superior performance with the highest Intersection over Union (IoU) score of 0.815. Our results in this study, coupled with the ablation assessment comparing scenarios with and without attention across various modalities, opens a promising path for enhancing the accuracy of flood analysis using remote sensing technology.

## I. Introduction

Floods, as the most frequent natural disaster, impact more people than any other natural hazard globally with its impacts on infrastructure and economy increasing each year due to climate change [1]. Furthermore, flooding disproportionately affect the poor, exacerbating poverty and hindering economic growth [2]. Historically, the primary means of understanding the impact of these hydrological events were through ground-based observations. However, these approaches were often limited in scope and lack the ability to provide real-time insights over large areas [3].

Remote sensing have made it possible to monitor vast and often inaccessible regions from space using earth observation, offering a comprehensive view of flood dynamics [4] [5] [6]. Earth observation can provide crucial information on the extent of flooding, the rate of water spread, potential future inundation areas, and the subsequent management recommendation [5].

Deep learning has been utilized for flood segmentation in recent years. Bai, et al. [4] leveraged both multispectral Sentinel-1 and Synthetical Aperture Radar (SAR) Sentinel-2 data and concatenated the extracted features to detect water during flood disasters using BASNet [7]. Muñoz et al. [8] took a similar approach, utilizing deep learning techniques to fuse together data from multispectral imagery, SAR, and digital elevation models (DEM) to map compound floods at both local and regional scales. However, all of these naïve concatenation approaches ignore the fact that different spectral imagery might complement or contrast each other depend on the input-output conditions [4][8]. To tackle this issue, Yadav et al. [5] proposed a deep attentive fusion network which allows the model to attentively choose which spectral image features are more useful to perform flood segmentation through a channel-wise attention network. Similarly, Oktay et al. [9] expanded on this concept with the Attention UNet, although the application was focused specifically on the medical image segmentation. Their work emphasized the importance of directing the model's attention to specific regions of interest, enhancing the accuracy of semantic segmentation [9] However, those approaches did not consider the correlative nature among different multispectral imagery, making the fusion approach less effective for multispectral semantic segmentation.

This paper proposes ProCANet, a novel semantic segmentation model for flood segmentation by leveraging both self and cross attention mechanisms, progressively applied to different size of intermediate features generated by pooling operation in a UNet like architecture. In particular, we utilized multispectral imagery (i.e. blue, green, red, and near infrared bands) and let our self and cross attention models to perform self-filtering and implicit cross-correlation among the multispectral features, generating feature combinations well suited for flood segmentation. We validated our approach by testing it with publicly available data (i.e., Sen1Floods11) and to our manually-digitized data, showing superior performance compared to the state-of-the-art semantic segmentation models.

## II. Proposed Methods

### A. Input Bands

The model that we proposed employed 2 multispectral images consisting of RGB (Red, Green, Blue) and NIR (Near Infra-Red). The fusion of RGB and NIR modalities offer rich representation of the scene, capturing both visible and non-visible spectrums. RGB images are in the visible spectrum and are commonly used to provide detailed information about the color and texture of features. On the other hand, NIR can be used to capture water-related features, such as moisture content and water boundaries, which might be invisible to the naked eye. The NIR spectrum has been extensively utilized in various studies to differentiate between water and non-water features. For instance, Wang et al. [10] utilized near-infrared spectroscopy to detect water content in fresh leaves, providing a basis for its potential in differentiating between water and non-water entities. In traditional remote sensing, the application of RGB and NIR bands from earth observation sensors is common place for mapping land cover [11], however, such band combinations are less common for semantic segmentation computer vision applications which were originally developed for RGB cameras. In our study we aimed to leverage the information in the NIR band by concatenating RGB with NIR as the input of the first encoder and utilizing just the NIR band as the input of the second encoder. Based on our experiment, this combination gave the best result in terms of accuracy, F1-score, and Intersection over Union (IoU) (see Section III).

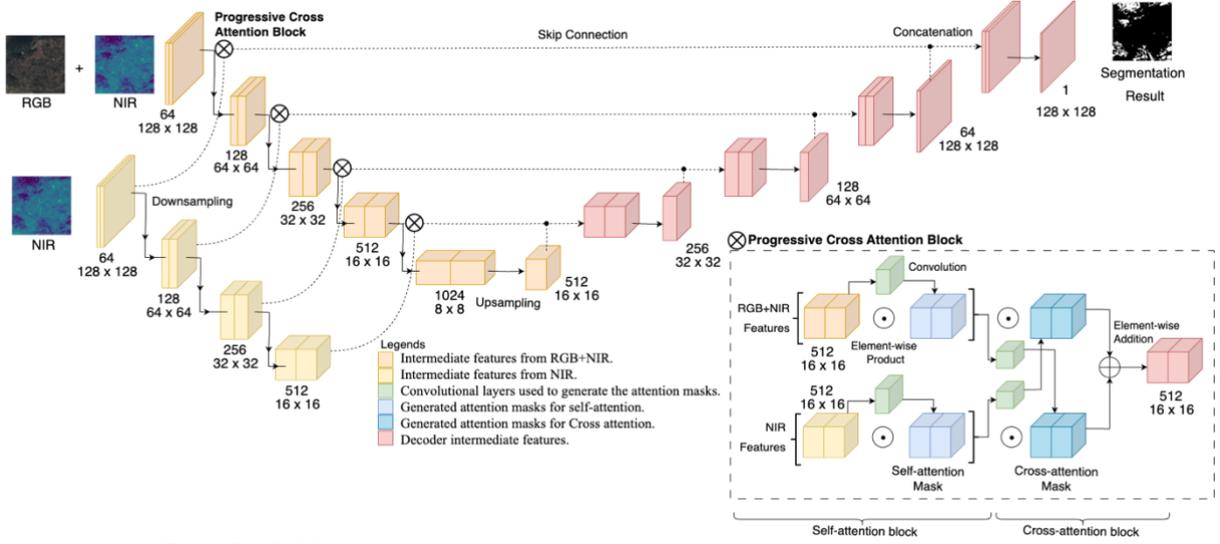

Fig. 1. ProCANet model's architecture which includes the progressive cross attention block.

*B. Deep Learning Architecture*

Our ProCANet model blends the advantages of UNet's [12] encoding and decoding architecture with the integration of our progressive cross-modal attention mechanisms. Designed to process 2 modalities, the model incorporates separate, yet intertwined encoder equipped with convolutional and pooling modules for each input modality. This allows for the extraction and manipulation of image details at various scales, effectively balancing the trade-off between contextual and locational information. Cross attention modules foster interactions between the modalities at each spatial scale, enabling an enriched exchange of spatial and spectral information. These attention mechanisms perform a two-fold task: amplifying modality-specific relevant features while suppressing the irrelevant ones (self-attention) and fostering the assimilation of insights across modalities (cross-attention). With a final convolutional layer mapping the resulting deep, cross-modality-fused feature maps to the desired output classes, the model delivers the final segmentation map, providing architecture for image segmentation tasks as can be seen in Fig. 1.

*C. Attention Architecture*

The attention architecture is designed to implement a progressive, multi-modal fusion approach, particularly focusing on the interaction between 2 modalities in respect of the channels. The term progressive refers to the nature of the attention mechanisms that are gradually applied to the intermediate 3D spatial features (or sometimes referred as the 'skip features') generated after every pooling operation in the encoder network. The details of the proposed progressive cross attention block, which is comprised of self-attention and cross-attention mechanisms, is described as follows.

*1) Self-attention Stage:* In the self-attention stage, we employ the features from each modality to generate the attention masks used to attend themselves, hence the term self-attention. Let a set of intermediate features generated by the encoder's convolutional layers are represented by $\{X_R^i, X_N^i\} \in \mathbb{R}^{w \times h \times c}$, where $X_R^i$ denotes features generated by encoder with RGB+NIR bands and $X_N^i$ denotes features generated by encoder with only NIR band. Note that $i, w, h,$ and $c$ are the $i$-th pooling layer, the width, the height, and the feature's channel (number of bands) respectively. To generate the self attention mask, $\{X_R^i, X_N^i\}$ is passed through a convolutional layer and a non-linear activation function described as follows:

$$a_R^i = \sigma\left(W_R^i(X_R^i)\right)$$
$$a_N^i = \sigma\left(W_N^i(X_N^i)\right)$$

In particular, the convolution operation involves the application of kernel with weight **W** which convolves across the input map, performing element-wise multiplication with the section of the input it is currently overlaying, and summing up all these multiplied values to obtain a single pixel in the output feature map. Specifically, both first and second inputs in our architecture are independently convolved using a 3x3 convolutional kernel with padding of 1. The convoluted feature maps are passed through a non-linear sigmoid activation function to generate the attention map. The sigmoid function ensures that the resultant attention map contains values in the [0, 1] range, facilitating a probabilistic interpretation of feature importance. The original intermediate features $\{X_R^i, X_N^i\}$ are then element-wise multiplied by the generated attention map $\{a_R^i, a_N^i\}$ to produce the self-attended feature map as follows:

$$\widehat{X}_R^i = X_R^i \odot a_R^i$$
$$\widehat{X}_N^i = X_N^i \odot a_N^i$$

Given the self-attention operations, the attended features $\{\widehat{X}_R^i, \widehat{X}_N^i\}$ will represent amplified relevant modality-specific features, while the irrelevant ones are suppressed.

*2) Cross-attention Stage:* The cross-attention mechanism is designed to combine information between different modalities, ensuring that the model not only recognizes intrinsic patterns within individual modalities but also captures and leverages the inter-modal dependencies. Given the attended features $\{\widehat{X}_R^i, \widehat{X}_N^i\}$, two respective cross attention masks are generated as follows:

$$a_{R \to N}^i = \sigma(W_{R \to N}^i(\widehat{X}_R^i))$$
$$a_{N \to R}^i = \sigma(W_{N \to R}^i(\widehat{X}_N^i))$$

In the subsequent cross-attention stage, the attended features are convoled with another 3x3 convolutional kernel, followed by a sigmoid activation function. The output of this operation is then used to modulate both previously obtained self-attended features through element-wise multiplication represented as below:

$$\widetilde{\mathbf{X}}_R^i = \widehat{\mathbf{X}}_R^i \odot a_{N \to R}^i$$
$$\widetilde{\mathbf{X}}_N^i = \widehat{\mathbf{X}}_N^i \odot a_{R \to N}^i$$

where $\{\widetilde{\mathbf{X}}_R^i, \widetilde{\mathbf{X}}_N^i\}$ are the generated cross attended features for both encoders in the specific *i*-th pooling operation.

*3) Cross Attended Fusion:* Finally, the cross-attended features from the first and second encoders are fused together through an element-wise addition operation, resulting in a composite feature map $\widetilde{\mathbf{X}}_{R+N}^i$ that incorporates both modalities as below:

$$\widetilde{\mathbf{X}}_{R+N}^i = \widetilde{\mathbf{X}}_R^i \oplus \widetilde{\mathbf{X}}_N^i$$

This cross-attention mechanism enables an intricate interaction between the two modalities, allowing the model to selectively emphasize and combine information, thereby enhancing the representational capacity of the network in multi-modal learning scenarios.

*D. Training Strategy*

To increase the amount of training data and to facilitate the extraction of more granular information, we systematically cut the original images into non-overlapping patches of $128 \times 128$ pixels. The selection of this specific patch size was the result of an iterative assessment, and represents a balance between the necessity for adequate local structural detail and the imperatives of computational efficiency and adequate training. Within the context of the training data loader, a method encompassing 64 patch steps was constructed to derive more data from the patching process. A filtering algorithm was also incorporated, serving to exclude patches with more than 50% of the pixel values equal to 255. This filtering mechanism was developed to ensure that the training set comprised patches containing substantial informative content, thus augmenting the model's capacity to discern land cover features. In contrast, the testing data was designed with an expanded 128 patch steps, making sure that there are no overlapping regions among the patches.

To train the model, we leverage a composite loss function that combines dice loss and soft binary cross entropy with logits. The dice loss is defined as follows:

$$\mathcal{L}_D = 1 - \frac{2 * \sum(y_{true} * y_{pred}) + \epsilon}{\sum y_{true} + \sum y_{pred} + \epsilon}$$

where $y_{true}$ is the ground truth, $y_{pred}$ is the predicted output and $\epsilon = 1e^{-7}$ is a smoothing factor to avoid division by zero. On the other hand, binary cross entropy is defined as follows

$$\mathcal{L}_{BCE} = -\frac{1}{N}\sum[y_{true} * \log(y_{pred}) + (1 - y_{true}) * \log(1 - y_{pred})]$$

where $y_{true}$ is the ground truth values, $y_{pred}$ is the predicted output from the model before the sigmoid activation (logits). Note that N is the total number of elements in $y_{true}$. The final objective function is a combination of the binary cross entropy and dice loss, described as follows:

$$\mathcal{L}_{final} = \mathcal{L}_{BCE} + \mathcal{L}_D$$

where $\mathcal{L}_{final}$ is the combined losses.

III. EXPERIMENTS AND DISCUSSIONS

*A. Dataset*

For the main experiments, including for training and testing, we utilized Sen1Floods11 dataset [13]. The dataset consists of Sentinel-1 and Sentinel-2 satellite imagery consisting of 455 total images where each of them are hand-labelled manually. Each of the image consists of 512 x 512 pixels. In our experiment, we specifically used multispectral Sentinel-2 for the purpose of training. The dataset was divided into two distinct subsets, with 65% allocated to the training set and the remaining 35% designated for validation.

To evaluate the model's generalization capability, we tested it using satellite imagery acquired from a new location with different image resolutions. We used PlanetScope Satellite imagery (planet.com) with a 5-meter resolution, which is 6 times higher than Sen1Floods11. The imagery was captured in the upper Citarum River Basin, Java, Indonesia, covering a total area of 6,112 km² as a single, large image (divided into 15 patches). Since ground truth flood segmentation was not available for this imagery, we manually mapped the flooded regions. This process involved using a spectral index (NDWI) for preliminary delineation and making manual corrections based on true-color images. Factors such as spatial arrangement, surrounding features, texture, color, tones, and patterns were considered to identify the extent of the flood. Due to the inability to accurately validate the manual delineation, we refer to these labels as a modified NDWI (pseudo ground truth).

*B. Training Details*

The optimization of the model's parameters is achieved by using the Adam optimizer with learning rate 0.0001. For controlling the learning rate, a Cosine Annealing scheduler with Warm Restarts is employed. The scheduler incorporates multiple restarts, where the learning rate is annealed following a cosine function between the initial learning rate and zero. Specifically, the training process utilizes a total of ten restarts, with each period doubling in length, enabling cyclical learning rate behavior. The training proceeds for a total of 25 epochs.

*C. Evaluation Metrics, Baselines and Ablation Study*

To evaluate the performance of ProCANet, we utilized common metrics for segmentation tasks namely accuracy, F1 score, and Intersection over Union (IoU). While F1 score complements accuracy as it shows harmonic mean between precision and recall, IoU is determined by aggregating the mean intersection and union over the entire dataset. Finally, in the ablation study we test the impact with and without attention and from different modalities (i.e., band combinations). For the quantitative comparison, we compared the accuracy of our proposed approach with the state-of-the-art semantic segmentation models (Table I). The competitors are UNet [12], PSPNet [14], LinkNet [15], MANet [16], PAN [17] and ConvNeXt V2 [18]. Note that MANet and PAN represents the model that incorporate attention to the model. We trained all these models using MobileNetV2 as the backbones and utilized R+G+B+NIR as the bands.

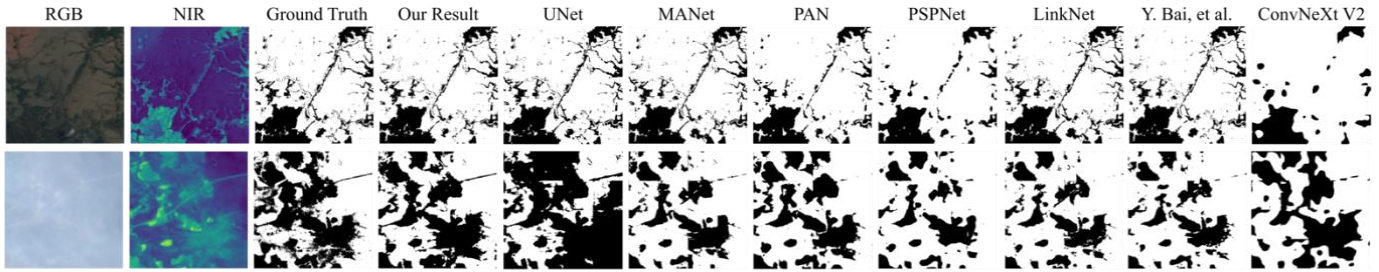

Fig. 2. Comparison of the flood segmentation outputs between our model and the state-of-the-art models in Sen1Floods11.

### D. Results

1) Evaluation on Sen1Floods11's test set

Fig. 2 represents the visualization of our model's segmentation in Sen1Floods11's test set compared to other state-of-the-art models. This qualitative assessment shows that our model segmentation output is nearly identical to the ground truth labels, indicating the effectiveness of our model in segmenting flooded areas in the images. The model was even able to accurately segment flooded areas in low quality images with blurry pixels likely caused by haze from clouds (row 2) which makes some of the other models struggle.

TABLE I Performance Comparison on Sen1Floods11 Test Set

| Model | Accuracy | F1-Score | IoU |
|---|---|---|---|
| UNet [12] | 0.9789 | 0.8844 | 0.791 |
| Y. Bai, et al. [4] | 0.9720 | 0.8380 | 0.722 |
| PSPNet [14] | 0.9702 | 0.8329 | 0.714 |
| LinkNet [15] | 0.9762 | 0.8681 | 0.767 |
| MANet [16] | 0.9779 | 0.878 | 0.783 |
| PAN [17] | 0.9736 | 0.8563 | 0.749 |
| ConvNeXt V2 [18] | 0.9660 | 0.8109 | 0.682 |
| **ProCANet (Ours)** | **0.9811** | **0.8982** | **0.815** |

TABLE I shows that our model achieved the highest accuracy score of 0.9811, F1-Score of 0.8982 and IoU score of 0.815, demonstrating superior segmentation performance. This underscores the efficacy of our progressive cross attention network compared to recent state-of-the art models like ConvNeXt V2 and other attention-based models such as MANet and PAN. The suboptimal performance from previous works might be due to their inability to fully leverage the correlative nature among different feature modalities. On the other hand, our progressive cross attention network performed self-filtering and implicit cross-correlation among the multispectral features in different scales, maximizing the complementary nature of different modalities.

2) Generalization test with Citarum PlanetScope imagery

The model's generalization performance on the Citarum river PlanetScope sample dataset is reflected in Fig. 3 and Table II. As shown in Fig. 3, despite not retraining the model on the PlanetScope imagery, which has a resolution six times higher than Sen1Floods11, the model was still able to produce reasonable predictions (clear and smooth delineation between inundated and non-inundated regions). The quantitative comparison with the modified NDWI (pseudo ground truth), indicated by an IoU of 0.659, as shown in TABLE II, further validates the model's generalization capabilities. Note that the lower IoU compared to the results from Sen1Floods11 is due to the differences in the approaches to generate the (pseudo) ground truth (the modified NDWI) in the dataset.

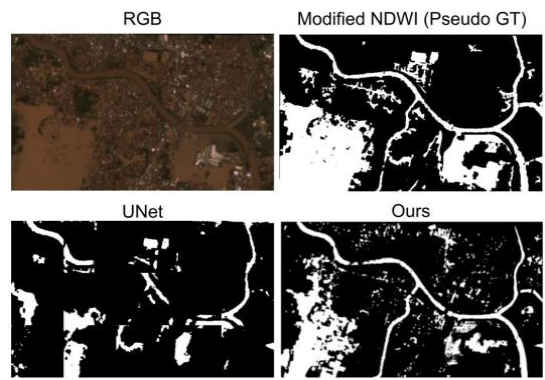

Fig. 3. Example results of the proposed model applied to our ground truth data for the Citarum river using high resolution 5 m imagery from PlanetScope.

TABLE II Quantitative Result on the PlanetScope Data Against a Modified NDWI (Pseudo GT)

| Model | Accuracy | F1 Score | IoU Score |
|---|---|---|---|
| UNet | 0.788 | 0.672 | 0.539 |
| **Ours** | 0.794 | 0.724 | 0.659 |

### E. Ablation study of attention mechanism and modalities

TABLE III compares the outcome between our proposed model with attention and without the attention mechanism. Note that we combine the two encoders of the model without attention by using element-wise addition. As we can see, our proposed model performs better than the model without attention, highlighting the importance of incorporating our progressive cross attention mechanism. To further understand the importance of the proposed progressive cross-attention mechanism, Fig. 4 illustrates the extracted features before (from the 1$^{st}$ and the 2$^{nd}$ encoders) and after the attention blocks. As we can see, the features generated after the attention blocks strongly highlight the edges in which the water and non-water area in the image is more visibly separated. NIR only modality, which is represented in the 2nd encoder, also strongly influences the attended features, highlighting the importance of NIR spectrum to delineate water and non-water features.

TABLE III Performance of The Proposed Model with and Without Attention Mechanism

| Second encoder | Attention mechanism | Accuracy | F1-Score | IoU |
|---|---|---|---|---|
| Yes | No | 0.980 | 0.891 | 0.804 |
| Yes | Yes | **0.981** | **0.898** | **0.815** |

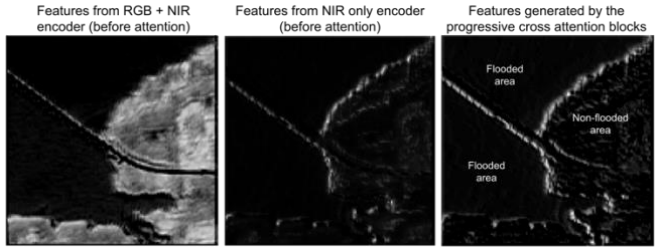

Fig. 4. Example of the extracted features before (in the 1st and 2nd encoder) and after the attention blocks.

TABLE IV shows the comparison of our experiment among modalities. The RGB modality with no encoder 2 represent the standard UNet model. The results highlight the importance of selecting appropriate modalities, with the combination of RGB + NIR and NIR resulting in a higher IoU. Incorporating additional NIR in first encoder (with RGB) slightly improves IoU, as it might help to compensate for any missing information in RGB encoder.

TABLE IV Performance of Our Model Using Different Modalities

| Encoder 1 | Encoder 2 | Accuracy | F1-Score | IoU |
|---|---|---|---|---|
| RGB | - | 0.948 | 0.653 | 0.483 |
| NIR | - | 0.961 | 0.776 | 0.634 |
| RGB, NIR | SAR | 0.972 | 0.838 | 0.722 |
| RGB, NIR | - | 0.980 | 0.891 | 0.804 |
| RGB, NDWI | NIR | 0.980 | 0.892 | 0.805 |
| RGB | NIR | **0.981** | 0.897 | 0.813 |
| RGB, NIR | NIR | **0.981** | **0.898** | **0.815** |

## IV. CONCLUSIONS

This study proposes a novel progressive cross attention network which utilizes both self and cross attention mechanism to produce the best multispectral feature combinations for flood segmentation using remote sensing data. Our proposed model has shown that it is capable of generating accurate flood segmentation with better results than state-of-the-art segmentation models applied to the Sen1Floods11 dataset and our Citarum river flood dataset. Exploration of different approaches such as utilizing other modalities and employing other fusion technique could be explored in future work.

## V. ACKNOWLEDGEMENTS

This work was funded by Monash Indonesia Research Innovation Seed Funding Grant (2022).